\documentclass[10pt,twocolumn,letterpaper]{article}

\usepackage{cvpr}
\usepackage{times}
\usepackage{epsfig}
\usepackage{graphicx}
\usepackage{amsmath}
\usepackage{amssymb}
\usepackage{comment}

\usepackage{color}
\usepackage{booktabs}
\usepackage{multirow}
\usepackage{array}
\usepackage{xcolor}
\usepackage{subcaption}
\usepackage{graphicx}
\usepackage{hhline}
\usepackage{makecell}
\usepackage{colortbl}
\usepackage[normalem]{ulem}
\usepackage{soul}
\usepackage{url}
\usepackage{pifont}% http://ctan.org/pkg/pifont
\newcommand{\cmark}{\ding{51}}%
\newcommand{\xmark}{\ding{55}}%
\usepackage[square,comma,numbers,sort]{natbib}
\usepackage[
  pagebackref=true,
  breaklinks=true,
  letterpaper=true,
  colorlinks,
  bookmarks=false,
  citecolor=red,
  linkcolor=blue]{hyperref}

\cvprfinalcopy % *** Uncomment this line for the final submission

\makeatletter
\DeclareRobustCommand\onedot{\futurelet\@let@token\@onedot}
\def\@onedot{\ifx\@let@token.\else.\null\fi\xspace}

\def\eg{\emph{e.g}\onedot}

\def\etal{\emph{et al}\onedot}

\makeatother
\newcommand{\thickhline}{%
    \noalign {\ifnum 0=`}\fi \hrule height 1pt
    \futurelet \reserved@a \@xhline
}

\def\cvprPaperID{****} % *** Enter the CVPR Paper ID here

\ifcvprfinal\fi
\begin{document}

%%%%%%%%% TITLE
\title{An Implementation of Faster RCNN with Study for Region Sampling}

\author{Xinlei Chen\\
Carnegie Mellon University\\
{\tt\small xinleic@cs.cmu.edu}
\and
Abhinav Gupta\\
Carnegie Mellon University\\
{\tt\small abhinavg@cs.cmu.edu}
}

\maketitle
%\thispagestyle{empty}

%%%%%%%%% ABSTRACT
\begin{abstract}
   We adapted the join-training scheme of Faster RCNN framework from Caffe to TensorFlow as a baseline implementation for object detection. Our code is made publicly available. This report documents the simplifications made to the original pipeline, with justifications from ablation analysis on both PASCAL VOC 2007 and COCO 2014. We further investigated the role of non-maximal suppression (NMS) in selecting regions-of-interest (RoIs) for region classification, and found that a biased sampling toward small regions helps performance and can achieve on-par mAP to NMS-based sampling when converged sufficiently.
\end{abstract}

%%%%%%%%% BODY TEXT
\newcolumntype{x}{>\small c}
\newcolumntype{L}[1]{>{\raggedright\let\newline\\\arraybackslash\hspace{0pt}}m{#1}}
\newcolumntype{C}[1]{>{\centering\let\newline\\\arraybackslash\hspace{0pt}}m{#1}}
\newcolumntype{R}[1]{>{\raggedleft\let\newline\\\arraybackslash\hspace{0pt}}m{#1}}
\begin{table*}[t]
\centering
\renewcommand{\arraystretch}{1.2}
\renewcommand{\tabcolsep}{1.2mm}
\caption[caption]{\label{tab:voc2007-tf}{\small{\bf VOC 2007 test} object detection average precision (\%). All use Faster RCNN on VGG16. Legend: \textbf{C}: Caffe implementation, results are generated with models trained with open-sourced code; \textbf{T}: our TensorFlow implementation; \textbf{P}: use RoI pooling; \textbf{A}: use \texttt{crop\_and\_resize}; \textbf{N}: images per batch; \textbf{R}: total RoIs for training the region classifier; \textbf{S}: keep small RoIs.}}
\vspace{-0.1in}
\resizebox{\linewidth}{!}{
\begin{tabular}{@{} C{0.3cm} C{.3cm} C{.3cm} C{.6cm} C{.3cm} !{\color{gray}\vrule} c !{\color{gray}\vrule} *{20}{x} @{}}
\Xhline{1pt}
 &  & \textbf{N} & \textbf{R} & \textbf{S} & mAP & aero      & bike      & bird      & boat      & bottle     & bus        & car        & cat        & chair      & cow        & table      & dog        & horse      & mbike      & persn     & plant      & sheep      & sofa       & train      & tv  \\
\Xhline{1pt}
\textbf{C} & \textbf{P} & 2 & 128 & \xmark & 70.0 & 68.7 & \textbf{79.2} & 67.6 & 54.1 & 52.3 & 75.8 & 79.8 & 84.3 & 50.1 & 78.3 & 65.1 & \textbf{82.2} & \textbf{84.8} & 72.9 & 76.0 & 44.9 & 70.9 & 63.3 & 76.1 & 72.6 \\
\textbf{C} & \textbf{P} & 1 & 128 & \xmark & 70.1 & 69.4 & 78.3 & 67.4 & 55.9 & 53.9 & \textbf{81.5} & 84.6 & \textbf{85.2} & 49.3 & 74.3 & 62.4 & 80.6 & 83.2 & 74.6 & 76.9 & 43.0 & \textbf{72.1} & 62.1 & 75.6 & 71.3 \\
\Xhline{.5pt}
\textbf{T} & \textbf{P} & 1 & 128 & \xmark & 69.8 & \textbf{71.9} & 76.4 & \textbf{67.8} & 54.8 & 51.9 & 80.1 & \textbf{85.4} & 83.8 & 46.9 & 73.6 & 65.7 & 79.8 & 80.2 & 75.4 & 76.2 & 41.4 & 70.5 & 65.0 & 75.7 & \textbf{73.5} \\
\textbf{T} & \textbf{A} & 1 & 128 & \xmark & 69.9 & 69.2 & 77.5 & 67.1 & 56.9 & 52.9 & 75.8 & \textbf{85.4} & 82.5 & 47.4 & 77.8 & 61.9 & 81.2 & 83.5 & \textbf{76.0} & 76.9 & 43.2 & 69.9 & 65.9 & 74.4 & 72.1 \\
\textbf{T} & \textbf{A} & 1 & 256 & \xmark & 70.7 & 69.4 & 78.9 & 66.8 & \textbf{57.4} & 55.7 & 75.2 & 85.1 & 83.0 & \textbf{50.6} & 80.0 & \textbf{67.6} & 81.4 & 80.4 & 74.5 & 77.1 & 42.8 & 71.1 & 66.3 & \textbf{77.1} & \textbf{73.5} \\
\textbf{T} & \textbf{A} & 1 & 256 & \cmark & \textbf{70.9} & 67.5 & 78.4 & 67.0 & 53.4 & \textbf{58.9} & 78.2 & 85.1 & 84.4 & 49.2 & \textbf{82.1} & 66.7 & 77.3 & 84.3 & 75.4 & \textbf{77.3} & \textbf{46.2} & 71.0 & \textbf{66.6} & 75.2 & \textbf{73.5} \\
\Xhline{1pt}
\end{tabular}
}
\vspace{-0.1in}
\end{table*}

\begin{table*}[t]
\centering
\renewcommand{\arraystretch}{1.1}
\renewcommand{\tabcolsep}{1.2mm}
\caption[caption]{\label{tab:coco-tf}{\small{\bf COCO 2014 minival} object detection average precision and recall (\%) with provided evaluation tool. For Caffe we use the released model. Legend same as in Table~\ref{tab:voc2007-tf}.}}
\vspace{-0.1in}
\resizebox{0.75\linewidth}{!}{
\begin{tabular}{@{} C{0.3cm} C{.3cm} C{.3cm} C{.6cm} C{.3cm} !{\color{gray}\vrule} c !{\color{gray}\vrule} *{11}{x} @{}}
\Xhline{1pt}
 &  & \textbf{N} & \textbf{R} & \textbf{S} & AP & AP-.5 & AP-.75 & AP-S & AP-M & AP-L & AR-1 & AR-10 & AR-100 & AR-S & AR-M & AR-L \\
\Xhline{1pt}
\textbf{C} & \textbf{P} & 2 & 128 & \xmark & 24.2 & 45.1 & 23.4 & 7.4 & 27.5 & \textbf{38.2} & 23.6 & 33.7 & 34.3 & 11.7 & 39.5 & \textbf{54.1} \\
\Xhline{0.5pt}
\textbf{T} & \textbf{A} & 1 & 128 & \xmark & 25.2 & 44.7 & 25.6 & 9.4 & 29.4 & 37.5 & 24.3 & 35.0 & 35.7 & 14.2 & 41.4 & 53.4\\
\textbf{T} & \textbf{A} & 1 & 256 & \xmark & 26.0 & 45.8 & 27.1 & 10.5 & \textbf{30.4} & 38.1 & 24.8 & 35.7 & 36.3 & 15.0 & \textbf{42.5} & 53.0\\
\textbf{T} & \textbf{A} & 1 & 512 & \xmark & 25.7 & 45.4 & 26.3 & 10.0 & 29.9 & 37.3 & 24.4 & 35.1 & 35.8 & 14.1 & 41.7 & 52.2\\
\Xhline{0.5pt}
\textbf{T} & \textbf{A} & 1 & 128 & \cmark & 25.4 & 45.6 & 25.4 & 11.0 & 29.2 & 37.0 & 24.6 & 35.8 & 36.5 & 16.8 & 41.5 & 52.8\\
\textbf{T} & \textbf{A} & 1 & 256 & \cmark & \textbf{26.5} & \textbf{46.7} & \textbf{27.2} & \textbf{11.8} & \textbf{30.4} & 37.5 & \textbf{24.9} & \textbf{36.3} & \textbf{37.1} & \textbf{17.3} & 42.1 & 52.4\\
\textbf{T} & \textbf{A} & 1 & 512 & \cmark & 26.1 & 46.4 & 26.3 & 11.6 & 29.9 & 36.6 & 24.7 & 35.8 & 36.5 & 16.4 & 41.8 & 51.9\\
\Xhline{1pt}
\end{tabular}
}
\vspace{-0.2in}
\end{table*}

\section{Baseline Faster RCNN with Simplification}

We adapted the join-training scheme of Faster RCNN detection framework\footnote{\url{https://github.com/rbgirshick/py-faster-rcnn}}~\cite{ren2015faster} from Caffe\footnote{\url{https://github.com/BVLC/caffe}} to TensorFlow\footnote{\url{https://github.com/tensorflow}} as a baseline implementation. Our code is made publicly available\footnote{\url{https://github.com/endernewton/tf-faster-rcnn}}. During the implementation process, several simplifications are made to the original pipeline, with observations from ablation analysis that they are are either not affecting or even potentially improving the performance. The ablation analysis has the following default setup:
\begin{description}
  \item[Base network.] Pre-trained VGG16~\cite{simonyan2014very}. The feature map from \texttt{conv5\_3} are used for region proposals and fed into region-of-interest (RoI) pooling.
  \item[Datasets.] Both PASCAL VOC 2007~\cite{everingham2010pascal} and COCO 2014~\cite{lin2014microsoft}. For VOC we use the \texttt{trainval} split for training, and \texttt{test} for evaluation. For COCO we use \texttt{train+valminusminival} and \texttt{minival}, same as the published model.
  \item[Training/Testing.] The default end-to-end, single-scale training/testing scheme is copied from the original implementation. Learning rate starts with $.001$ and is reduced after $50k$/$350k$ iterations. Training finishes at $70k$/$490k$ iterations. Following COCO challenge requirements, for each testing image, the detection pipeline provides at most $100$ detection results.
  \item[Evaluation.] We use evaluation toolkits provided by the respective dataset. The metrics are based on detection average precision/recall.
\end{description}

The first notable change follows Huang \etal~\cite{google}. Instead of using the RoI pooling layer, we use the \texttt{crop\_and\_resize} operator, which crops and resizes feature maps to $14\times 14$, and then max-pool them to $7\times 7$ to match the input size of \texttt{fc6}.

Second, we do not aggregate gradients from $N=2$ images and $R=128$ regions~\cite{girshick2015fast}, instead we simply sample $R=256$ regions from $N=1$ images during a single forward-backward pass. Gradient accumulation across multiple batches is slow, and requires extra operators in TensorFlow. Note that $R$ is the number of regions sampled for training the region classifier, for training region proposal network (RPN) we still use the default $256$ regions.

Third, the original Faster RCNN removes small proposals (less than $16$ pixels in height or width in the original scale). We find this step redundant, hurting the performance especially for small objects.

Other minor changes that does not seem to affect the performance include: 1) double the learning rate for bias; 2) stop weight decay on bias; 3) remove aspect-ratio grouping (introduced to save memory); 4) exclude ground-truth bounding boxes in the RoIs during training, since they are not accessible during testing and can bias the input distribution for region classification.

For ablation analysis results on VOC 2007, please check at Table~\ref{tab:voc2007-tf}. Performance-wise, our implementation is in general on par with the original Caffe implementation. The \texttt{crop\_and\_resize} pooling appears to have a slight advantage over RoI pooling.

We further test the pipeline on COCO, see Table~\ref{tab:coco-tf}. We fix $N=1$ and only use \texttt{crop\_and\_resize} pooling, which in general gives better average recall than RoI pooling. Keeping the small region proposals also gives consistent boost on small objects. Overall our baseline implementation gives better AP ($+4\%$) and AR ($+5\%$) for small objects. As we vary $R$, we find $256$ gives a good trade-off with the default training scheme, as further increasing $R$ causes potential over-fitting. 

\subsection{Training/Testing Speed}

Ideally, our training procedure can almost cut the total time in half since gradient is only accumulated over $N=1$ image. However, the increased batch size $R=256$ and the use of \texttt{crop\_and\_resize} pooling slow each iteration a bit. Adding the underlying TensorFlow overhead, the average speed for a COCO net on a Titan X (non Pascal) GPU for training is $400$ms per iteration, whereas for testing it is $160$ms per image in our experimental environment.

\begin{table*}[t]
\centering
\renewcommand{\arraystretch}{1.2}
\renewcommand{\tabcolsep}{1.2mm}
\caption[caption]{\label{tab:voc2007}{\small{\bf VOC 2007 test} object detection average precision (\%). Analysis of different region sampling schemes for train/test combinations. Baseline (first row) uses NMS for both training and testing. Please refer to Section~\ref{sec:inv} for the detailed meaning of ALL, PRE, POW and TOP, none of which is based on NMS.}}
\vspace{-0.1in}
\resizebox{\linewidth}{!}{
\begin{tabular}{@{} C{1.cm} C{1.cm} !{\color{gray}\vrule} c !{\color{gray}\vrule} *{20}{x} @{}}
\Xhline{1pt}
\textbf{Train} & \textbf{Test} & mAP & aero      & bike      & bird      & boat      & bottle     & bus        & car        & cat        & chair      & cow        & table      & dog        & horse      & mbike      & persn     & plant      & sheep      & sofa       & train      & tv  \\
\Xhline{1pt}
\small{NMS} & \small{NMS} & 70.9 & 67.5 & 78.4 & 67.0 & 53.4 & \textbf{58.9} & 78.2 & 85.1 & 84.4 & 49.2 & \textbf{82.1} & 66.7 & 77.3 & 84.3 & 75.4 & 77.3 & \textbf{46.2} & \textbf{71.0} & 66.6 & 75.2 & \textbf{73.5} \\
\Xhline{0.5pt}
\small{ALL} & \small{TOP} & 70.4 & \textbf{73.9} & 77.7 & 67.0 & 56.6 & 47.7 & 80.3 & 83.8 & 83.8 & 48.0 & 77.9 & \textbf{68.6} & \textbf{80.8} & 84.0 & \textbf{76.5} & 75.7 & 41.6 & 69.2 & 66.6 & 77.6 & 70.3 \\
\small{PRE} & \small{TOP} & 71.1 & 72.7 & \textbf{79.0} & 67.3 & \textbf{58.8} & 53.3 & \textbf{80.9} & 85.2 & \textbf{84.8} & \textbf{50.6} & 80.3 & 66.4 & 80.1 & 83.5 & 74.2 & \textbf{77.6} & 44.3 & 69.7 & 65.7 & 76.9 & 70.9\\
\small{POW} & \small{TOP} & 71.0 & \textbf{73.9} & 78.5 & 67.1 & 57.7 & 53.1 & 80.1 & \textbf{85.8} & 83.6 & 50.0 & 80.0 & 65.6 & 80.6 & 80.5 & 75.4 & 76.8 & 44.4 & 70.6 & 66.0 & \textbf{78.3} & 72.6\\
\Xhline{0.5pt}
\small{NMS} & \small{TOP} & \textbf{71.2} & 67.6 & 78.9 & \textbf{67.6} & 55.2 & 56.9 & 78.8 & 85.2 & 83.9 & 49.8 & 81.9 & 65.5 & 80.1 & \textbf{84.4} & 75.7 & \textbf{77.6} & 45.3 & 70.8 & \textbf{66.9} & 78.2 & 72.9\\
\Xhline{1pt}
\end{tabular}
}
\vspace{-0.1in}
\end{table*}

\section{A Study of Region Sampling\label{sec:inv}}

\begin{table*}[t]
\centering
\renewcommand{\arraystretch}{1.1}
\renewcommand{\tabcolsep}{1.2mm}
\caption[caption]{\label{tab:coco}{\small{\bf COCO 2014 minival} object detection average precision and recall (\%) with provided evaluation tool. Baseline (first row) uses NMS for both training and testing. Please refer to Section~\ref{sec:inv} for the detailed meaning of ALL, PRE, POW and TOP, none of which is based on NMS. stepsize is the number of train iterations before the learning rate is reduced; and itersize is the total number of iterations. }}
\vspace{-0.1in}
\resizebox{0.82\linewidth}{!}{
\begin{tabular}{@{} C{1.cm} C{1.cm} !{\color{gray}\vrule} C{1.3cm} C{1.3cm} !{\color{gray}\vrule} c !{\color{gray}\vrule} *{11}{x} @{}}
\Xhline{1pt}
\textbf{Train} & \textbf{Test} & stepsize & itersize & AP & AP-.5 & AP-.75 & AP-S & AP-M & AP-L & AR-1 & AR-10 & AR-100 & AR-S & AR-M & AR-L \\
\Xhline{1pt}
\small{NMS} & \small{NMS} & $350k$ & $490k$ & 26.5 & 46.7 & 27.2 & 11.8 & 30.4 & 37.5 & 24.9 & 36.3 & 37.1 & 17.3 & 42.1 & 52.4 \\
\Xhline{0.5pt}
\small{ALL} & \small{TOP} & $350k$ & $490k$ & 23.2 & 41.2 & 23.7 & 7.1 & 24.1 & 36.9 & 23.0 & 32.9 & 33.5 & 12.1 & 36.5 & 52.8 \\
\small{PRE} & \small{TOP} & $350k$ & $490k$ & 25.1 & 44.1 & 25.7 & 9.0 & 27.4 & 38.8 & 24.4 & 35.1 & 35.7 & 14.1 & 39.5 & 55.0 \\
\small{POW} & \small{TOP} & $350k$ & $490k$ & 25.2 & 44.6 & 25.6 & 9.6 & 28.3 & 37.6 & 24.4 & 35.5 & 36.4 & 14.9 & 40.5 & 55.5 \\
\Xhline{0.5pt}
\small{NMS} & \small{TOP} & $350k$ & $490k$ & 26.9 & 47.0 & 27.7 & \textbf{12.0} & 31.0 & 38.9 & 25.3 & 37.2 & 38.1 & 17.6 & 43.1 & 54.0 \\
\Xhline{0.5pt}
\small{ALL} & \small{TOP} & $600k$ & $790k$ & 25.0 & 43.5 & 25.4 & 7.8 & 26.1 & 39.5 & 24.2 & 34.4 & 35.1 & 13.1 & 38.2 & 55.1 \\
\small{PRE} & \small{TOP} & $600k$ & $790k$ & 26.6 & 45.7 & 27.7 & 9.8 & 29.3 & 41.5 & 25.3 & 36.5 & 37.3 & 15.3 & 41.9 & 56.0 \\
\small{POW} & \small{TOP} & $600k$ & $790k$ & 26.9 & 46.4 & 28.2 & 10.8 & 29.8 & 40.8 & 25.4 & 36.8 & 37.6 & 16.2 & 42.1 & 56.6 \\
\Xhline{0.5pt}
\small{NMS} & \small{NMS} & $600k$ & $790k$ & 27.9 & 48.2 & 29.0 & 11.8 & 31.8 & 40.3 & 26.0 & 37.5 & 38.3 & 17.6 & 43.4 & 55.4 \\
\small{NMS} & \small{TOP} & $600k$ & $790k$ & \textbf{28.3} & \textbf{48.7} & \textbf{29.5} & 11.8 & \textbf{32.5} & \textbf{41.9} & \textbf{26.2} & \textbf{38.3} & \textbf{39.2} & \textbf{18.0} & \textbf{44.3} & \textbf{56.7} \\
\Xhline{1pt}
\end{tabular}
}
\vspace{-0.2in}
\end{table*}

We also investigated how the distribution of the region proposals fed into region classification can influence the training/testing process. In the original Faster RCNN, several steps are taken to select a set of regions: 
\begin{itemize}
  \item First, take the top $K$ regions according to RPN score.
  \item Then, non-maximal suppression (NMS) with overlapping ratio of $0.7$ is applied to perform de-duplication.
  \item Third, top $k$ regions are selected as RoIs.
\end{itemize}
For training, $K=12000$ and $k=2000$ are used, and later $R$ regions are sampled for training the region classifier with pre-defined positive/negative ratio ($0.25/0.75$); for testing $K=6000$ and $k=300$ are used. We refer to this default setting as \textbf{NMS}.

In Ren \etal~\cite{ren2015faster}, a comparable mean average precision (mAP) can be achieved when the top-ranked $K=6000$ proposals are directly selected without NMS during \emph{testing}. This suggests that NMS can be removed at the cost of evaluating more RoIs. However, it is less clear whether NMS de-duplication is necessary during \emph{training}. On a related note, NMS is believed to be crucial for selecting hard examples for Fast RCNN~\cite{ohem_cvpr16}. Therefore, we want to check if it is also true for Faster RCNN in the joint-training setting.  

Our first alternative (\textbf{ALL}) works by simply feeding all top $K$ regions for positive/negative sampling without NMS. While this alternative appears to optimize the same objective function as the one with NMS, there is a subtle difference: NMS implicitly biases the sampling procedure toward smaller regions. Intuitively, it is more likely for large regions to overlap than small regions, so large regions have a higher chance to be suppressed. A proper bias in sampling is known to help at least converge networks more quickly~\cite{bansal2016pixelnet} and is actually also used in Faster RCNN: a fixed positive/negative ratio to avoid always learning on negative patches. To this end, we add two more alternatives for comparison. The first one (\textbf{PRE}) computes the final ratio of a pre-trained Faster RCNN model that uses NMS, and samples regions based on this final ratio. The second one (\textbf{POW}) simply fits the sampling ratio to the power law: $r(s)=s^{-\gamma}$ where $r$ is ratio, $s$ is scale, and $\gamma$ is a constant factor (set as $1.$). While PRE still depends on a trained model with NMS, POW does not require NMS at all. To fit the target distribution, we keep all regions of the scale with the highest ratio in the distribution, and randomly select regions of other scales according to the relative ratio. \eg, if the distribution is $(0.4,0.2,0.2)$ for scales $(8, 16, 32)$, then all the scale-$8$ regions are kept, and $50\%$ of the other two scales are later sampled. Note that for both of them we set $k=6000$ ($k$ is functioning as $K$) during training, since roughly half the regions are already thrown away.

Following Ren \etal~\cite{ren2015faster}, we simply select top $K$ proposals for evaluation directly. With little or no harm on precision but direct benefit on recall, mAP generally increases as $K$ gets larger. We set $K=5000$ trading off speed and performance. This testing scheme is referred as \textbf{TOP}.

We begin by showing results on VOC 2007 in Table~\ref{tab:voc2007}. As can be seen, apart from ALL, other schemes with biased sampling all achieve the same level of mAP (around $71\%$). We also include results (last row) that uses NMS during training but switches to TOP for testing. Somewhat to our surprise, it achieves better performance. In fact, we find this advantage of TOP over NMS consistently exists when $K$ is sufficiently large.

A more thorough set of experiments were conducted on COCO, which are summarized in Table~\ref{tab:coco}. Similar to VOC, we find biased sampling (NMS, PRE and POW) in general gives better results than uniform sampling (ALL). In particular, with $490k$ iterations of training, NMS is able to offer a performance similar to PRE/POW after $790k$ iterations. Out of curiosity, we also checked the model trained with $790k$ NMS iterations, which is able to converge to a better AP ($28.3$ on \texttt{minival}) with the TOP testing scheme. We did notice that with more iterations, the gap between NMS and POW narrows down from $1.7$ ($490k$) to $1.4$ ($790k$), indicating the latter ones may catch up eventually. The difference to VOC suggests that $490k$ iterations are not sufficient to fully converge on COCO. Extra experiments with longer training iterations are needed for a more conclusive note.

{\small
\bibliographystyle{ieee}
% \bibliography{negation}

}

\end{document}